\def\year{2021}\relax
\documentclass[letterpaper]{article} 
\usepackage{aaai21}  
\usepackage{times}  
\usepackage{helvet} 
\usepackage{courier}  
\usepackage[hyphens]{url}  
\usepackage{graphicx} 
\usepackage{natbib}  
\usepackage{caption} 
\usepackage{multirow}
\usepackage{bm}
\usepackage{booktabs}
\usepackage{subfigure}
\usepackage{amsfonts}
\usepackage{color}
\usepackage[switch]{lineno}

\title{Reinforced Medical Report Generation with\\  X-Linear Attention and Repetition Penalty}
\author{
	
		Wenting Xu,\textsuperscript{\rm 1 \rm 2}
	Chang Qi, \textsuperscript{\rm 1 \rm 2}
	Zhenghua Xu \textsuperscript{\rm 1 \rm 2*} 
	Thomas Lukasiewicz \textsuperscript{\rm 3} \\

}
\affiliations{
	
	\textsuperscript{\rm 1} State Key Laboatory of Reliablity and Intelligence of Electrical Equipment, Hebei University of Technology, China. \\
		\textsuperscript{\rm 2} Tianjin Key Laboratory of Bioelectromagnetic Technology and Intelligent Health, Hebei University of Technology, China. \\
		\textsuperscript{\rm 3}Department of Computer Science, University of Oxford, United Kingdom. \\

}

\begin{document}

	\maketitle
	
	\begin{abstract}
	To reduce doctors’ workload, deep-learning-based automatic medical report generation has recently attracted more and more research efforts, where attention mechanisms and reinforcement learning are integrated with the classic encoder-decoder architecture to enhance the performance of deep models.
	However, these state-of-the-art solutions mainly suffer from two shortcomings: (i) their attention mechanisms cannot utilize high-order feature interactions, and (ii) due to the use of TF-IDF-based reward functions, these methods are fragile with generating repeated terms. 
    Therefore, in this work, we propose a reinforced medical report generation solution with x-linear attention and repetition penalty mechanisms (ReMRG-XR) to overcome these problems.
    Specifically, x-linear attention modules are used to explore high-order feature interactions and achieve multi-modal reasoning, while repetition penalty is used to apply penalties to repeated terms during the model's training process.
    Extensive experimental studies have been conducted on two public datasets, and the results show that ReMRG-XR greatly outperforms the state-of-the-art baselines in terms of all metrics.

		
	\end{abstract}
	
	\section{Introduction}
	\noindent 

    \noindent Nowadays, medical imaging is the most commonly used medical examination method in disease diagnosis, and medical imaging reports are paragraph-based documents issued by radiologists after examinations. These reports describe the important medical findings observed on the corresponding medical images and emphasize the abnormalities, and the sizes and locations of detected lesions. 
   However, due to the increasing number of patients and the shortage of experienced radiologists, a radiologist may have to conduct dozens or sometimes even hundreds of medical imaging examinations and then write the same number of reports every day, which makes the radiologists overloaded and may lead to misdiagnoses.
    Therefore, it is a compelling demand to find promising ways to generate medical reports automatically.
	
%
	
	Existing deep-learning-based medical report generation solutions mainly adopt the encoder-decoder architecture~\cite{Wang_2018,jing-etal-2018-automatic,xue2018multimodal}, where deep convolutional neural networks (CNNs) are applied to encode the input medical images, and recurrent neural networks (RNNs), e.g., based on long short-term memory (LSTM) units, are then used as a decoder to generate medical reports automatically.
    However, these models are maximum likelihood-based autoregressive models, which inevitably suffer from the problems of exposure bias and inconsistent training goals~\cite{li2018hybrid}.
	
	Therefore, in order to enhance the deep model's performance in medical report generation, most existing works have integrated various attention mechanisms into the encoder-decoder architecture ~\cite{Wang_2018,jing-etal-2018-automatic,xue2018multimodal}. However, visual and/or semantic attention cannot exploit the channel-wise information and high-order feature interactions of medical images and texts to obtain fine-grained visual and semantic information, which thus limits the performance of these models. 
	
	Alternatively, some recent works~\cite{10.1007/978-3-030-32692-0_77,liu2019clinically} have introduced reinforcement learning (RL) into this task, where a term frequency inverse document frequency (TF-IDF) based metric, named CIDEr, is used as the reward. However,  CIDEr mainly focuses on term-level optimization, and ignores the importance of diversity in sentence-level; so,  these methods inevitably suffer from the problem of repeated terms, which thus weakens the coherency and readability of the generated medical reports.
	
	
	
	Consequently, in this paper, to overcome the above problems, we propose a \textbf{Re}inforced \textbf{M}edical \textbf{R}eport \textbf{G}eneration solution with \textbf{X}-linear attention and \textbf{R}epetition penalty mechanisms (abbreviated by ReMRG-XR). Intuitively, we believe that the utilization of high-order feature interactions will strengthen the model's capability in intra-modal and inter-modal reasoning, which will enhance the model performance in generating accurate medical reports. Therefore, to improve the accuracy of generated reports, we propose to use x-linear attention modules, which are a stack of bi-linear pooling-based attention blocks, to calculate the outer product between two feature vectors and enable channel-wise attention. This consequently enhances the model's capability in exploring high-order feature interactions and multi-modal reasoning between medical images and their corresponding reports during both encoding and decoding procedures, resulting in a better medical report generation performance.
	Furthermore, we also believe that if appropriate penalties are applied to repeated terms during the model's training process, it will be helpful for the model to generate more diverse terms and increase the coherency and readability of the generated reports.
	Therefore, we further propose to integrate a repetition penalty mechanism into the training process of reinforcement learning.
	Overall, the contributions of this paper are as follows:
	\begin{itemize}
		\item We identify a main limitation of existing medical report generation solutions in utilizing high-order feature interactions, and propose to use x-linear attention modules to achieve a more accurate medical report generation. 
		
		\item To avoid generating repeated terms, repetition penalty is used in the training process of reinforcement learning to enhance the coherency and readability of the generated medical reports.
		
		
		\item We conducted extensive experiments on two publicly available medical image report datasets.
		Experimental results show that our proposed ReMRG-XR model greatly outperforms the state-of-the-art baselines in terms of all metrics. we have also conducted ablation studies to prove that both the x-linear attention and the repetition penalty mechanism are effective and essential for the model to achieve superior performance.
		
	\end{itemize}

\section{Methods}

\begin{figure*}[!tb]
	\centering {\includegraphics[width=0.98\textwidth]{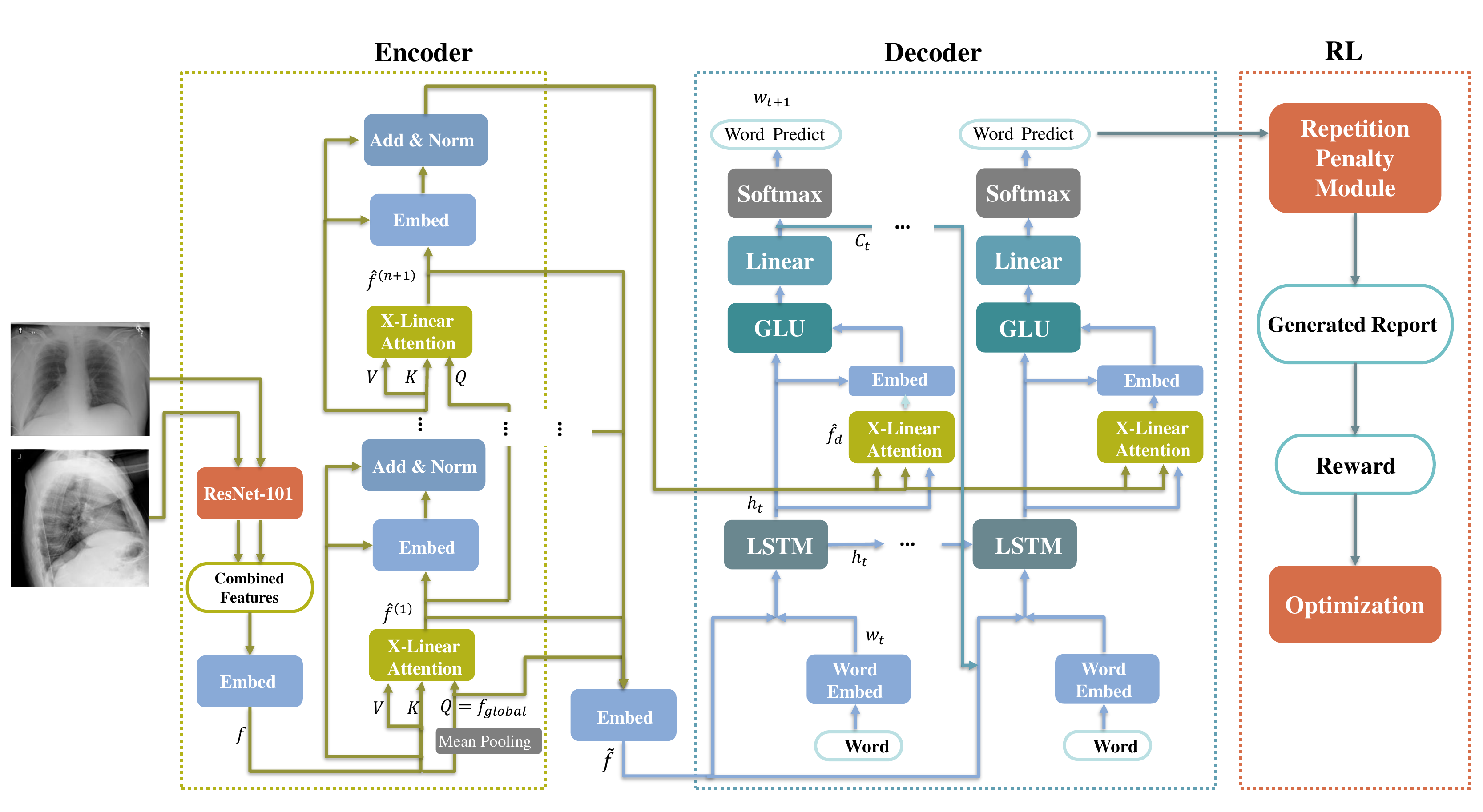}}
	\caption{The architecture of our proposed ReMRG-XR network. 
	}\vspace*{-1ex}
	\label{fig:framework}
\end{figure*}


We propose a reinforced medical report generation method with x-linear attention and repetition penalty (ReMRG-XR). Intuitively, we believe that the use of high-order feature interactions will strengthen the model's capacity in single- and multi-modal reasoning, which will enhance the model's performance in terms of accuracy of the generated reports. Besides, we consider that the combination of the integrated penalty with the repetition term will produce much more diverse sentences, which will increase the coherency and readability of the generated diagnosis reports.  

Specifically, as shown in Figure~\ref{fig:framework}, to improve the clinical accuracy of the generated reports, 
we take the global visual features extracted from medical images $f_{global}$ as input \textbf{Q} and the regional features $f$ as input \textbf{V} and \textbf{K}. A stack of x-linear attention blocks are used to calculate the outer product between two feature vectors and enable  channel-wise attention through the squeeze-excitation operation. As such, high-order intra-modal feature interactions are explored during the encoder procedure. 
After that, the embedded attended features $\hat{f}$ are combined and sent into LSTMs during the decoder procedure. The output hidden state of the LSTMs together with the final attended visual features are then sent into an x-linear attention block to explore multi-modal feature interactions, which is later used for word prediction.
After pretraining with such an encoder-decoder architecture for epochs, reinforcement learning is used to boost the performance, during which we employ the SCST algorithm and set CIDEr as the reward. To generate more readable descriptions, we use a repetition penalty in SCST to increase the readability of the generated reports.

\subsection{Reinforcement Learning with Diversity}
The reinforcement learning algorithm commonly used in existing medical report generation 
\cite{10.1007/978-3-030-32692-0_77,liu2019clinically} is the self-critical sequence training (SCST) algorithm \cite{rennie2017self}, which directly optimizes the automatic natural language generation (NLG) metrics. Since most of the existing works employing SCST take CIDEr  \cite{vedantam2015cider} as the reward, which focuses on phrase-level accuracy, ignoring the global fluency and diversity, we consider diversity-based~SCST with repetition penalty integrated.

SCST adopts a policy gradient method to optimize a non-differentiable metric such as CIDEr. In order to normalize the reward and reduce the variance during training, SCST utilizes the REINFORCE algorithm with a baseline, which is obtained from the inference reward by greedy search. The goal is to minimize the negative expected reward.
The final gradient of the optimization object is:

\begin{equation}\label{Eq:Eq14}\small
\frac{\partial{L(\theta)}}{\partial{s_t}} = (r(w^s) - r(\hat{w}))\bigtriangledown_{\theta}\log p_{\theta}(w^s | x), 
\end{equation}
where $w^s  = (w_1^s, . . . ,w_T^s )$ is a Monte-Carlo sample from~$p_{\theta}$, $p_{\theta}$ is our generation model,  $r(w^s)$ is the current reward, and $r(\hat{w})$ is the reward obtained by the inference algorithm.

\subsubsection{Repetition Penalty}
In order to decrease the probability of repeated terms generated in the reports,
we thus introduce a penalty that would constraint the probabilities of words resulting in repeated trigrams. Specifically, we update the log-probability of the output word by subtracting a value proportional to the number of times the trigram has been generated.

\begin{equation}
p_w = p_w - n_w \times \alpha, 
\end{equation}
where $p_w$ is the log-probability of the word $w$, $n_w$ is the number of times the trigram has generated the word $w$, and $\alpha$ is a hyperparameter. In our work, we set $\alpha$ to 2. We employ this update mechanism during our greedy search process in SCST in order to generate more diverse paragraphs, thus avoiding the repetition problem.

\section{Experiments}

\begin{table*}[ht] 
	
	\centering   
	\small
	\begin{tabular}{l|l|ccccccc}
		\hline  
		Dataset & Model & CIDEr & ROUGE-L  & METEOR & BLEU-1 & BLEU-2 & BLEU-3 & BLEU-4   \\ \hline 
		\multirow{6}{*}{\scriptsize{\textbf{IU X-Ray}}} 
		& top-down* & 0.206 & 0.334 & 0.144 & 0.279 & 0.178 & 0.119 & 0.079  \\ 
		&MRMA* &  0.325 & 0.309 & 0.163 & 0.382 & 0.252 & 0.173 & 0.120  \\ 
		&CACG & -- & \textbf{0.359} &--  & 0.369 & 0.246 & 0.171 & 0.115   \\ 
		\cline{2-9}   
		& MRG-X & 0.393 & 0.300 & 0.148 & 0.303 & 0.196 & 0.136 & 0.099  \\ 
		& ReMRG-X & \textbf{0.451} &0.346 & 0.180 & 0.391 & 0.262 & 0.184 & 0.131  \\ 
		& ReMRG-XR & 0.433 & \underline{0.357} & \textbf{0.184} & \textbf{0.402} & \textbf{0.276} &\textbf{ 0.196 }& \textbf{0.140}   \\ 
		\midrule 
		\multirow{6}{*}{\scriptsize{\textbf{MIMIC-CXR}}}  
		& top-down* & 0.359& 0.319& 0.134 & 0.233 & 0.159 & 0.119 & 0.093  \\ 
		&MRMA* & 0.324 & 0.330 & 0.157 & 0.361 & 0.244 & 0.182 & 0.141  \\
		&CACG & -- &0.307 & -- & 0.352 & 0.223 & 0.153 & 0.104  \\ 
		\cline{2-9}  
		& MRG-X & 0.383&0.323 & 0.156 & 0.345 & 0.231 & 0.170 & 0.131  \\ 
		& ReMRG-X & \textbf{0.415}&0.342 & 0.176 & 0.407 & 0.276 & 0.203 & 0.154  \\ 
		& ReMRG-XR & 0.411 & \textbf{0.342 }& \textbf{0.179} & \textbf{0.412} & \textbf{0.279} & \textbf{0.206 }& \textbf{0.157}  \\ 
		\hline 
	\end{tabular} 
	\caption{Automatic natural language evaluation on IU X-Ray (upper part) and MIMIC-CXR (lower part), compared with top-down~\protect\cite{anderson2018bottom}, 
		MRMA \cite{xue2018multimodal}, and 
		CACG \cite{liu2019clinically}, 
		* indicates our re-implementation. 	
	}
	\label{Table:report-generation} 
\end{table*}

\subsection{Implementation Details}
We use  ResNet-101 pre-trained on ImageNet \cite{deng2009imagenet} to extract the region-level features, which are from the last convolutional layer. As both views of medical images are sent into the model simultaneously, two original 2048-dimensional region features are concatenated to a 4096-dimensional vector, which is later transformed into the visual embedding of size 1024.
Next, four stacks of x-linear attention blocks are used to explore the high-order intra-modal interactions.
During \textit{Decode}, we set the hidden layer size and word embedding dimension to 1024.
As most reports are within a fixed length, we set the max generation length of IU X-Ray and MIMIC-CXR to 114 and 184, respectively.

In training, we first pre-train the model with cross-entropy loss for 60 epochs with a batch size of 8 with NVIDIA RTX 2080 GPUs. Setting the base learning rate to  0.0001, we utilize the Noam decay strategy with 10,000 warmup steps with Adam \cite{kingma2014adam} as the optimizer.
In the reinforcement learning stage, we select CIDEr as our training reward. We set the base learning rate to 0.00001 and decay with CosineAnnealing with a period of 15 epochs. We also set the maximum iteration to  60 epochs and the batch size to  2. We use beam search with a beam size of 2 for training.

\begin{figure*}[!tb]
	\centering {\includegraphics[width=0.98\textwidth]{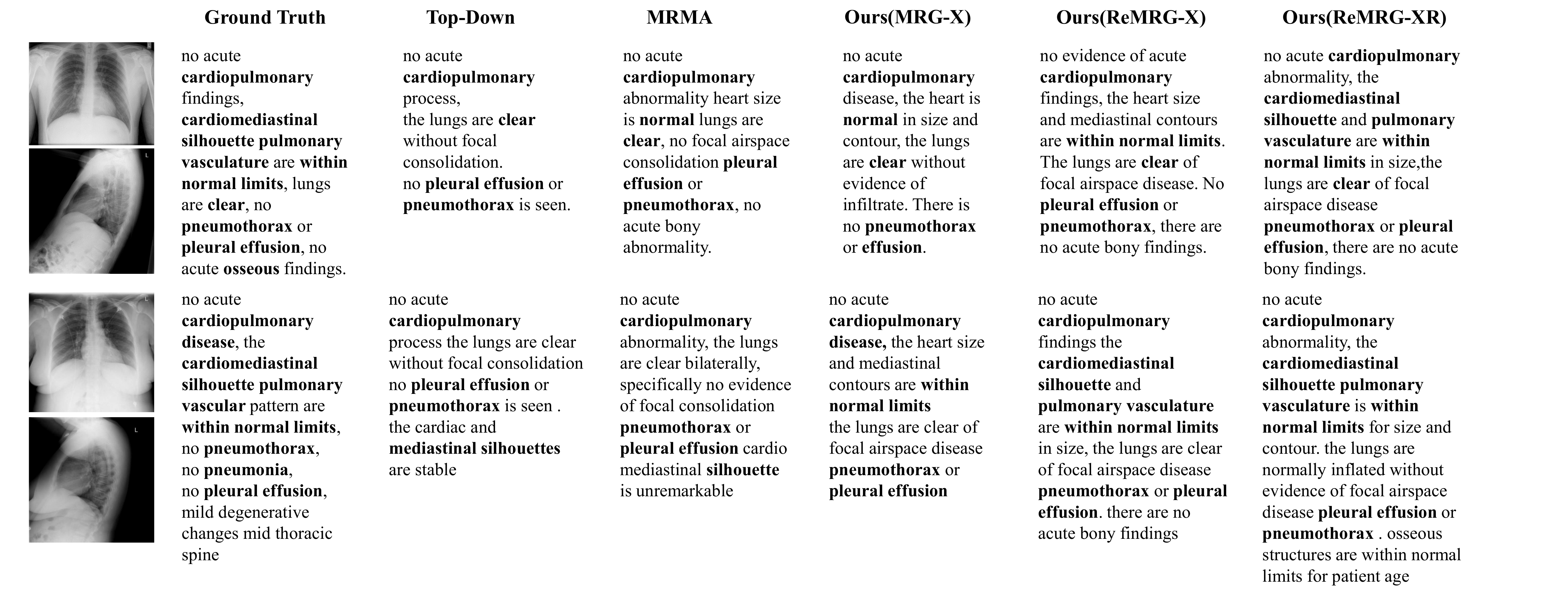}}
	\caption{Examples of reports generated by ReMRG-XP, ReMRG-X, MRG-X, MRMA, and Top-Down.}
	\label{fig:sample}
\end{figure*}

\subsubsection{Main Results.}
Table~\ref{Table:report-generation} shows the experimental results of the proposed ReMRG-XP, two simplified versions (ReMRG-X and MRG-X), and three baselines in terms of seven natural language generation metrics, where two of the baselines are re-implemented by ourselves (denoted by *). In addition, Figure~\ref{fig:sample} exhibits some examples of reports generated by these methods.
Generally, ReMRG-XP greatly outperforms all state-of-the-art baselines in all metrics  in Table~\ref{Table:report-generation}, and as in Figure~\ref{fig:sample}, it also generates more coherent and accurate report than the baselines. This finding proves the effectiveness of ReMRG-XP.
Furthermore, The performance of ReMRG-X is generally worse than ReMRG-XP, except for CIDEr. This is because ReMRG-X uses CIDEr as reward which is fragile with generating repeated terms, while repetition penalty is used in ReMRG-XP to overcome this problem, making it capable of generating more coherent and readable descriptions. This observation is also well supported by the report examples in Figure~\ref{fig:sample}. 
Comparing to ReMRG-X and ReMRG-XP, the performance of MRG-X is not very satisfactory because reinforcement learning is not utilized in MRG-X. However, MRG-X still significantly outperforms the top-down baseline, proving the effectiveness of using x-linear attention to obtain higher order features.

\section{Related Work}

Most medical report generation approaches adopt encoder-decoder-based frameworks, which utilize CNNs as an encoder to extract image features and feed them into an RNN-based decoder to generate sequences. Recent works have been trying to explore more interactions between images and sentences via attention mechanisms  \cite{10.1007/978-3-030-26763-6_66,Wang_2018,jing-etal-2018-automatic,xue2018multimodal}. \citet{jing-etal-2018-automatic} propose a co-attention mechanism that builds attention distributions separately for visual and semantic domains and ignores multi-modal interactions. \citet{xue2018multimodal} develop a sentence-level attention mechanism to explore multi-modal interactions, which compute the attention distribution over visual regions according to sentence-level semantic features. 
However, all these methods solely explore first-order feature interactions; differently, in this work, both intra-model and inter-modal high-order feature interactions are explored by x-linear attention, and reinforcement learning is further used to enhance the model's performance. 


With  rapid advances in reinforcement learning,
many efforts have used reinforcement learning for automatic medical report generation to boost the performance \cite{10.1007/978-3-030-32692-0_77,liu2019clinically,li2018hybrid}. Specifically, \citet{li2018hybrid} propose a hybrid retrieval-based model with reinforcement learning to determine whether to generate a sentence via template retrieval or LSTMs. Subsequently,  \citet{10.1007/978-3-030-32692-0_77} and  \citet{liu2019clinically} adopt reinforcement learning to directly optimize the text generation.
Compared to existing reinforcement learning solutions, the proposed ReMRG-XR have two advantages: (i) it uses x-linear attention blocks to enhance the model's capability in exploring high-order feature interactions and multi-modal reasoning, and (ii) a repetition penalty is introduced to make the model generate more diverse and coherent long sentences.

\section{Conclusion and Future Work}

In this paper, we have proposed to use x-linear attention, which integrates bi-linear pooling to explore high-order feature interactions for intra-modal and inter-modal reasoning, for medical report generation. We have also used reinforcement learning with diversity to generate long paragraphs. We have conducted extensive experiments on two publicly available datasets, IU X-Ray and MIMIC-CXR, which have demonstrated the superior performance of our proposed approach to achieve state-of-the-art results.

\bibliography{report}
	
\end{document}